\def\year{2020}\relax
\documentclass[letterpaper]{article} 
\usepackage{aaai20}  
\usepackage{times}  
\usepackage{helvet} 
\usepackage{courier}  
\usepackage[hyphens]{url}  
\usepackage{graphicx} 
\usepackage{graphicx}  
\usepackage{makecell}
\usepackage{algorithm}
\usepackage{algorithmic}
\usepackage{subcaption}
\usepackage{verbatim}
\usepackage{tikz}
\usepackage{pgfplots}
\pgfplotsset{compat=1.14}
\usepackage{pgfplotstable}
\usepackage{pgffor,pgfmath}
\usetikzlibrary{calc}
\usetikzlibrary{arrows.meta}
\usepackage{multirow}
\usepackage{amsmath}
\usepackage{amssymb}
\usepackage{booktabs}
\usepackage{mathtools}
\usepackage{xcolor}
\usepackage{mathtools}
\newcommand{\cyan}[1]{{\color{cyan}#1}}
\newcommand{\red}[1]{{\color{red}#1}}

\usepackage{amsmath,amsfonts,bm}

\def\eqref#1{equation~\ref{#1}}

\def\1{\bm{1}}

\def\rs{{\textnormal{s}}}

\def\rv{{\textnormal{v}}}

\def\rmA{{\mathbf{A}}}

\def\rmE{{\mathbf{E}}}

\def\rmR{{\mathbf{R}}}

\def\rmW{{\mathbf{W}}}

\def\vx{{\bm{x}}}
\def\vy{{\bm{y}}}
\def\vz{{\bm{z}}}

\def\evx{{x}}
\def\evy{{y}}

\DeclareMathAlphabet{\mathsfit}{\encodingdefault}{\sfdefault}{m}{sl}
\SetMathAlphabet{\mathsfit}{bold}{\encodingdefault}{\sfdefault}{bx}{n}

\newcommand{\pdata}{p_{\rm{data}}}

\newcommand{\E}{\mathbb{E}}

\DeclareMathOperator*{\argmax}{argmax}
\DeclareMathOperator*{\argmin}{argmin}

\newcommand\eoi{\langle END\rangle}
\newcommand\cls{\langle CLS\rangle}
\newcommand\masked{\langle m\rangle}
\def\insertp#1#2#3{q(#1|#2\downarrow#3)}
\def\insertptok#1#2#3{\hat{q}(#1|#2\downarrow#3)}

\def\estinsertp#1#2#3{q_\theta(#1|#2\downarrow#3)}
\def\estinsertptok#1#2#3{\hat{q}_\theta(#1|#2\downarrow#3)}
\def\estinsertpcond#1#2#3#4{q_\theta(#1|#2\downarrow#3#4)}

\definecolor{lightblue}{RGB}{207,226,243}
\definecolor{lightgreen}{RGB}{217,234,211}
\definecolor{lightyellow}{RGB}{255,229,153}
\definecolor{sepia}{rgb}{0.44, 0.26, 0.08}
\definecolor{lightpurple}{RGB}{217,210,233}

\usepackage{ifthen}
\newboolean{revising}
\setboolean{revising}{true}
\ifthenelse{\boolean{revising}}
{
\newcommand{\peter}[1]{\textbf{\textcolor{red}{peter: #1}}}
\newcommand{\yssays}[1]{\textbf{\textcolor{purple}{ys: #1}}}

} {
\newcommand{\peter}[1]{}
\newcommand{\yssays}[1]{}

}

\nocopyright

\title{XL-Editor: Post-editing Sentences with XLNet
}
\author{
Yong-Siang Shih, Wei-Cheng Chang, Yiming Yang \\ 
Carnegie Mellon University\\
\{yongsias,wchang2,yiming\}@cs.cmu.edu
}
 
\begin{document}
\frenchspacing

\maketitle

\begin{abstract}
While neural sequence generation models achieve initial success for many NLP applications, the canonical decoding procedure with left-to-right generation order (i.e., autoregressive) in one-pass can not reflect the true nature of human revising a sentence to obtain a refined result.
In this work, we propose XL-Editor, a novel training framework that enables state-of-the-art generalized autoregressive pretraining methods, XLNet specifically, to revise a given sentence by the variable-length insertion probability.
Concretely, XL-Editor can (1) estimate the probability of inserting a variable-length sequence into a specific position of a given sentence; (2) execute post-editing operations such as insertion, deletion, and replacement based on the estimated variable-length insertion probability; (3) complement existing sequence-to-sequence models to refine the generated sequences.
Empirically, we first demonstrate better post-editing capabilities of XL-Editor over XLNet on the text insertion and deletion tasks, which validates the effectiveness of our proposed framework.
Furthermore, we extend XL-Editor to the unpaired text style transfer task, where transferring the target style onto a given sentence can be naturally viewed as post-editing the sentence into the target style. XL-Editor achieves significant improvement in style transfer accuracy and also maintains coherent semantic of the original sentence, showing the broad applicability of our method.

\end{abstract}

\section{Introduction}

Neural sequence generation is an essential component of modern deep learning models for applications such as machine translation \cite{sutskever2014sequence}, image captioning \cite{vinyals2015show}, text summarization \cite{nallapati2016abstractive}, and  speech synthesis \cite{oord2016wavenet}.
The most popular models often rely on the left-to-right autoregressive factorization of probability estimation, and decode/generate the output sequence token by token in a fixed order. By contrast, humans are more flexible in text writing, and are capable of making multiple revisions to the same text on arbitrary positions to obtain a refined result.

Very recently, the NLP community is witnessing groundbreaking progress in unsupervised pretraining models, where XLNet \cite{yang2019xlnet} stands as one of the most advanced development that significantly outperforms many competitors such as ELMo \cite{peters2018deep}, GPT-2 \cite{radford2019language} and BERT \cite{devlin-etal-2019-bert}.
In the heart of XLNet is the generalized autoregressive pretraining objective that optimizes the language modeling likelihood under all permutation factorization orders, which in expectation better captures the bidirectional context for each token position compared to BERT.

While these pretraining models reach state-of-the-art performance in many discriminative downstream tasks including sentence classification, question answering and sequence tagging, the text generation ability of these pretraining models remain under explored. For example, initial investigation showed the generation quality of BERT is not satisfactory~\cite{wang2019bert,mansimov2019generalized}.
More importantly, both BERT and XLNet implicitly requires prior knowledge of the sequence length to estimate the generated sequence probability given the context, which makes it difficult to sample variable-length sequences for post-editing generation.

Motivated by the pros and cons of XLNet for text generation, in this work, we propose XL-Editor, a novel training framework that extends XLNet to revise a given sentence via the generalized insertion operation.
\begin{itemize}
    \item Firstly, to estimate any variable-length subsequence given the context, XL-Editor incorporates a novel insertion-based relative positional encoding scheme to learn the variable-length insertion probability estimation efficiently.
    This overcomes the computation bottleneck existed in earlier insertion-based models~\cite{pmlr-v97-stern19a} as well as the limitation caused by the implicit length assumption in XLNet.
    \item Secondly, by solving variable-length insertion probability estimation, XL-Editor endows three post-editing operations (i.e., insertion, deletion, and replacement), which enables revision of a given sentence for higher text generation quality.
    \item Thirdly, our training framework can leverage the power of pretrained XLNet. By enabling XLNet with post-editing operations, we unlock the applicability of XLNet for wider range of tasks.
\end{itemize}

Empirically, we demonstrate the effectiveness of XL-Editor in two set of experiments.
\begin{itemize}
    \item Firstly, we devise three simple tasks including Locate (which position to insert), Text Infilling (what tokens to be inserted), and Text Deletion (what tokens to be deleted) to examine the post-editing capabilities. The proposed XL-Editor achieve better performance compared to the vanilla XLNet baselines.
    \item Secondly, we demonstrate the potential of post-editing on the unpaired text style transfer problem, where transferring the target style onto a given sentence can be naturally viewed as post-editing the sentence into the target style.
    We train conditional version of the probability estimation based on different styles and make our XL-Editor execute post-editing operations on the system outputs from the state-of-the-art methods.  We show that significant improvement in style accuracy can be achieved with little sacrifice in content preservation.
\end{itemize}

\section{Background and Related Work \label{sec:background}}
In the heart of XL-Editor model is to estimate the variable-length insertion probability for post-editing process. We begin with the objective of estimating the variable-length insertion probability, connected it with existing sequence modeling techniques, and point out their weakness in estimating such probabilities.
Finally, we briefly summarize some related works in the post-editing literature.

\subsection{Variable-length Insertion Probability Estimation}

Given two sequences $\vx^{(l)}$ and $ \vx^{(r)}$ as the left and the right contexts, respectively, we aim to estimate the probability of inserting a certain sequence $\vy$ between $\vx^{(l)}$ and $ \vx^{(r)}$, where $\vy$ can have arbitrary length. We denote such probability as $\insertp{\vy}{\vx^{(l)}}{\vx^{(r)}}$, and factorize it in the following manner,
\begin{align*}
    \insertp{\vy}{\vx^{(l)}}{\vx^{(r)}}= &\prod_{t=1}^{|\vy|} \insertptok{y_t}{\vx^{(l)} \oplus \vy_{<t}}{\vx^{(r)}} \\
    &\times \insertptok{\eoi}{\vx^{(l)}\oplus \vy}{\vx^{(r)}},
\end{align*}
where $\oplus$ denotes the concatenation operator, the special token, $\eoi$, denotes the end of insertion, and $\hat{q}(\cdot)$ denotes the probability of inserting a single token.
Note that this is similar to the autoregressive modeling methods but with additional left context $\vx^{(l)}$ and right context $\vx^{(r)}$.

Let $\vx = (\evx_1, \evx_2, \cdots, \evx_T)$ be a sequence of length $T$, $\evx_i$ be the $i$-th token in $\vx$, and $\vx_{i:j}$ be $(\evx_i, \evx_{i+1}, \cdots, \evx_{j})$, a subsequence of $\vx$, whose length is exactly $j-i+1$. We could denote the probability of inserting $\vy$ between the $i$-th and the $(i+1)$-th tokens of $\vx$ as $\insertp{\vy}{\vx_{1:i}}{\vx_{i+1:T}}$.

\subsection{Existing Sequence Modeling Techniques}

\subsubsection{Masked Sequence Modeling}
Given a masked sequence $\hat{\vx}$ and denote the masked tokens as $\vy$, masked sequence modeling methods aim to reconstruct the masked tokens $\vy$ from $\hat{\vx}$. The objective being optimized can be denoted as $\max_\theta p_\theta(\vy|\hat{\vx})$.
BERT \cite{devlin-etal-2019-bert} is one of the most prominent examples that utilizes masked sequences. While BERT achieves strong results in various discriminative downstream tasks, estimating variable-length insertion probabilities with masked sequence modeling remains challenging. Firstly, the probability estimation is only defined when the number of masked tokens in $\vy$ is exactly the same as the number of masking tokens in $\hat{x}$, and thus it's unsuitable for a variable-length $\vy$. Secondly, BERT assumed all tokens in $\vy$ are independent when making the prediction, which limits its modeling capacity.
\citeauthor{maskgan18}~\shortcite{maskgan18} proposed a masked sequence modeling method that does not have the independence assumption. However, their method still requires a predetermined length before making a prediction.

\subsubsection{Permutation Sequence Modeling} \citeauthor{yang2019xlnet}~\shortcite{yang2019xlnet} proposed XLNet to estimate the probability of a sequence based on any factorization orders.
In particular, given a sequence $\vx$, XLNet can model the probability of $ p_{\theta}(\vx_{i:j}|\vx_{1:i-1},\vx_{j+1:T}) $, where $T$ is the length of $\vx$ and $\vx_{i:j}$ is any subsequence of $\vx$. While it's tempting to utilize the modeling capacity of XLNet to estimate variable-length insertion probability, this is actually problematic because the probability estimation of XLNet also has an implicit assumption on the length of the subsequence being estimated just like masked sequence modeling methods if both the left and the right contexts are non-empty. The implicit length assumption is encoded by the relative positional encoding, which conveys the fact that the distance between the left context $\vx_{1:i-1}$ and right context $\vx_{j+1:T}$ is exactly $j-i+2$.

\subsubsection{Insertion-based Sequence Modeling} The Insertion Transformer~\cite{pmlr-v97-stern19a,chan2019kermit} directly models insertion operations with a joint probability over where to insert and which token to be inserted. 
Although Insertion Transformer offers to predict the insertion position, it's unclear what's the best insertion order to be used for training the model to make such predictions.
One advantage of Insertion Transformer and other non-autoreressive methods \cite{DBLP:conf/nips/SternSU18,gu2018nonautoregressive} is that they allow parallel decoding in multiple positions, which could potentially shorten the decoding time with some sacrifice in generation quality.

While our method may also generalize to multiple insertion positions, in this work we only focus on one insertion position at a time. 
On the other hand, our method has two major advantages over their model. Firstly, our method utilizes the variable-length insertion probability estimation for executing post-editing operations such as deletion and replacement, while their model does not offer a clear way to execute such operations. Due to their joint probability formulation, estimation for the variable-length insertion probability requires a marginalization over all possible insertion orders, which might be intractable when then sequence being estimated is too long. Secondly, their method requires re-computation of all hidden vectors for every new token inserted, which creates computation bottleneck in their training procedure. Our implementation, which will be discussed later,  does not suffer from such issues.

\subsubsection{Sequence-to-Sequence Modeling} Traditional autoregressive sequence modeling cannot make use of both the left and the right context simultaneously. The insertion probability estimation, nevertheless, can be formulated as a sequence-to-sequence (Seq2Seq) problem by treating both the left and right context as the source sequence while treating the inserted sequence as the target sequence. However, directly applying existing Seq2Seq models usually makes it difficult to encode the relative positional relationships between tokens in a consistent manner.
For example, 
\citeauthor{ippolito-etal-2019-unsupervised}~\shortcite{ippolito-etal-2019-unsupervised} utilized the transformer architecture for story infilling, while  \citeauthor{zhu2019text}~\shortcite{zhu2019text} leveraged self-attention to tackle the text infilling problem. Although \citeauthor{zhu2019text}~\shortcite{zhu2019text} presented their model as a purely self-attention method, they still make the model attend to the context separately from the inserted sequence, making their method more similar to the standard transformer with tied weights between the encoder and the decoder. In both of these works, the model treats the left context and the already inserted tokens differently when making the prediction on a token being inserted, which hinders their ability to estimate $q(\cdot)$ faithfully. In addition, their works only focused on generating the inserted sequence, while the aim of our method is to utilize the probability estimations to enable post-editing operations.

\subsection{Related Work in Post-editing}
Recently, there has been increasing interest in alleviating the restriction imposed by the fixed-order generation process. For example, it has been proposed to apply imitation learning and reinforcement learning for learning post-editing policies \cite{gu2019levenshtein,WuRLS19}. In addition, several works have investigated iterative refinement for machine translation systems~\cite{novak2016iterative,NIPS2017_6775,grangier-auli-2018-quickedit,Freitag2019APEAS}. However, these methods either require complicated training procedure or the availability of parallel corpus, which limits their applicability.

\section{XL-Editor \label{sec:xl_editor}}
In this section, we introduce our solution to tackle the difficulties of estimating variable-length insertion probabilities. First, we propose an insertion-based relative positional encoding scheme to enhance XLNet, allowing it to estimate probabilities without length assumption. Next, we discuss how to execute text editing with our method, and compare XL-Editor against XLNet on several simple post-editing tasks to validate our proposed enhancement.

\subsection{Insertion-based Relative Positional Encoding}

The main reason that XLNet struggles in estimating the probability for an arbitrary inserted sequence $\vy$ lies in the fact that its predictions have implicit length assumption. Specifically, the attention score between the $i$-th position and $j$-th position in a sequence is computed as,

\begin{align*}
	\rmA_{i,j}^\text{rel}
	&= {\rmE_{i}^\top \rmW_q^\top \rmW_{k,E} \rmE_{j} }
	+ {\rmE_{i}^\top \rmW_q^\top \rmW_{k,R} \cyan{\rmR_{i-j}} } \\
	&+ {u^\top \rmW_{k,E} \rmE_{j}}
	+ {v^\top \rmW_{k,R} \cyan{\rmR_{i-j}}},
\end{align*}
where $\rmE_{i}$ denotes the hidden vector at $i$-th position, $\rmW_{k,E}, \rmW_{k,R}, \rmW_{q}$ denote three learnable weight matrices, $u$ and $v$ denote two learnable weight vectors, and ${\rmR_{i-j}}$ denotes a sinusoid encoding vector as defined in \citeauthor{vaswani2017attention}\shortcite{vaswani2017attention}, where the positional parameter
is set as $(i-j)$. In particular, the relative positional encoding vector $\rmR_{i-j}$ being used conveys the relative positional relationship between the two positions, $i$ and $j$.

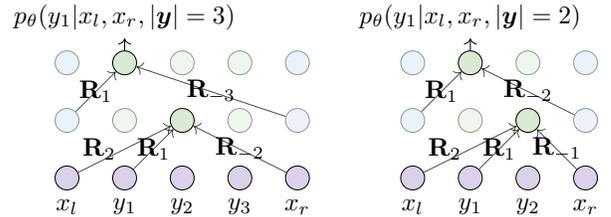
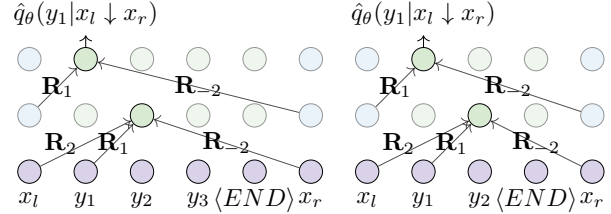
\begin{figure}[ht]
  \centering
  \begin{subfigure}{0.99\columnwidth}
  \centering
  \resizebox{.99\columnwidth}!{ 
  \begin{tikzpicture}[scale=0.8,every node/.style={scale=1}]
   \node[circle,draw,fill={lightpurple}] (xl1) at (1,0) {};
   \node at ($(xl1) + (0, -0.5cm)$) {$\evx_{l}$};
   
   \node[circle,draw,fill={lightpurple}] (y1) at (2,0) {};
   \node at ($(y1) + (0, -0.5cm)$) {$\evy_{1}$};
   \node[circle,draw,fill={lightpurple}] (y2) at (3,0) {};
   \node at ($(y2) + (0, -0.5cm)$) {$\evy_{2}$};
   \node[circle,draw,fill={lightpurple}] (y3) at (4,0) {};
   \node at ($(y3) + (0, -0.5cm)$) {$\evy_{3}$};
   
   \node[circle,draw,fill={lightpurple}] (xr0) at (5,0) {};
   \node at ($(xr0) + (0, -0.5cm)$) {$\evx_{r}$};
   
   \node[circle,draw,fill={lightblue}, opacity=0.4] (xl1_1) at (1,1) {};
   \node[circle,draw,fill={lightgreen}, opacity=0.4] (y1_1)  at (2,1) {};
   \node[circle,draw,fill={lightgreen}] (y2_1)  at (3,1) {};
   \node[circle,draw,fill={lightgreen}, opacity=0.4] (y3_1)  at (4,1) {};
   \node[circle,draw,fill={lightblue}, opacity=0.4] (xr0_1) at (5,1) {};
   \draw[<-,draw opacity=.7] (y2_1) -- node[left] {$\rmR_2$} ++ (xl1);
   \draw[<-,draw opacity=.7] (y2_1) -- node {$\rmR_1$} ++ (y1);
   \draw[<-,draw opacity=.7] (y2_1) -- node {$\rmR_{-2}$} ++ (xr0);
   
   \node[circle,draw,fill={lightblue}, opacity=0.4] (xl1_2) at (1,2) {};
   \node[circle,draw,fill={lightgreen}] (y1_2)  at (2,2) {};
   \node[circle,draw,fill={lightgreen}, opacity=0.4] (y2_2)  at (3,2) {};
   \node[circle,draw,fill={lightgreen}, opacity=0.4] (y3_2)  at (4,2) {};
   \node[circle,draw,fill={lightblue}, opacity=0.4] (xr0_2) at (5,2) {};
   \draw[<-,draw opacity=.7] (y1_2) -- node {$\rmR_1$} ++ (xl1_1);
   \draw[<-,draw opacity=.7] (y1_2) -- node {$\rmR_{-3}$} ++ (xr0_1);
   
   \node (pd) at ($(y1_2) + (0, +0.8cm)$) {$p_{\theta}(\evy_{1}|\evx_l, \evx_r, |\vy|=3)$};
   
   \draw[->] (y1_2) -- (pd);
   
   \node[circle,draw,fill={lightpurple}] (r_xl1) at (7,0) {};
   \node at ($(r_xl1) + (0, -0.5cm)$) {$\evx_l$};
   
   \node[circle,draw,fill={lightpurple}] (r_y1) at (8,0) {};
   \node at ($(r_y1) + (0, -0.5cm)$) {$\evy_{1}$};
   \node[circle,draw,fill={lightpurple}] (r_y2) at (9,0) {};
   \node at ($(r_y2) + (0, -0.5cm)$) {$\evy_{2}$};
   
   \node[circle,draw,fill={lightpurple}] (r_xr0) at (10,0) {};
   \node at ($(r_xr0) + (0, -0.5cm)$) {$\evx_r$};
   
   \node[circle,draw,fill={lightblue}, opacity=0.4] (r_xl1_1) at (7,1) {};
   \node[circle,draw,fill={lightgreen}, opacity=0.4] (r_y1_1)  at (8,1) {};
   \node[circle,draw,fill={lightgreen}] (r_y2_1)  at (9,1) {};
   \node[circle,draw,fill={lightblue}, opacity=0.4] (r_xr0_1) at (10,1) {};
   \draw[<-,draw opacity=.7] (r_y2_1) -- node[left] {$\rmR_2$} ++ (r_xl1);
   \draw[<-,draw opacity=.7] (r_y2_1) -- node {$\rmR_1$} ++ (r_y1);
   \draw[<-,draw opacity=.7] (r_y2_1) -- node {$\rmR_{-1}$} ++ (r_xr0);
   
   \node[circle,draw,fill={lightblue}, opacity=0.4] (r_xl1_2) at (7,2) {};
   \node[circle,draw,fill={lightgreen}] (r_y1_2)  at (8,2) {};
   \node[circle,draw,fill={lightgreen}, opacity=0.4] (r_y2_2)  at (9,2) {};
   \node[circle,draw,fill={lightblue}, opacity=0.4] (r_xr0_2) at (10,2) {};
   \draw[<-,draw opacity=.7] (r_y1_2) -- node {$\rmR_1$} ++ (r_xl1_1);
   \draw[<-,draw opacity=.7] (r_y1_2) -- node {$\rmR_{-2}$} ++ (r_xr0_1);
   
   \node (r_pd) at ($(r_y1_2) + (0, +0.8cm)$) {$p_{\theta}(\evy_{1}|\evx_l,\evx_r, |\vy|=2)$};
   \draw[->] (r_y1_2) -- (r_pd);
\end{tikzpicture}
  }
  \caption{\label{fig:xlnet_attn} Attention in XLNet, the query stream (green nodes) are computed differently depending on assumption of $|\vy|$.}
  \end{subfigure}
  \begin{subfigure}{0.99\columnwidth}
  \centering
  \resizebox{.99\columnwidth}!{ 
  \begin{tikzpicture}[scale=0.8,every node/.style={scale=1}]
   \node[circle,draw,fill={lightpurple}] (xl1) at (1,0) {};
   \node at ($(xl1) + (0, -0.5cm)$) {$\evx_l$};
   
   \node[circle,draw,fill={lightpurple}] (y1) at (2,0) {};
   \node at ($(y1) + (0, -0.5cm)$) {$\evy_{1}$};
   \node[circle,draw,fill={lightpurple}] (y2) at (3,0) {};
   \node at ($(y2) + (0, -0.5cm)$) {$\evy_{2}$};
   \node[circle,draw,fill={lightpurple}] (y3) at (4,0) {};
   \node at ($(y3) + (0, -0.5cm)$) {$\evy_{3}$};
   \node[circle,draw,fill={lightpurple}] (eoi) at (5,0) {};
   \node at ($(eoi) + (0, -0.5cm)$) {${ \eoi}$};
   
   \node[circle,draw,fill={lightpurple}] (xr0) at (6,0) {};
   \node at ($(xr0) + (0, -0.5cm)$) {$\evx_r$};
   
   \node[circle,draw,fill={lightblue}, opacity=0.4] (xl1_1) at (1,1) {};
   \node[circle,draw,fill={lightgreen}, opacity=0.4] (y1_1)  at (2,1) {};
   \node[circle,draw,fill={lightgreen}] (y2_1)  at (3,1) {};
   \node[circle,draw,fill={lightgreen}, opacity=0.4] (y3_1)  at (4,1) {};
   \node[circle,draw,fill={lightgreen}, opacity=0.4] (eoi_1)  at (5,1) {};
   \node[circle,draw,fill={lightblue}, opacity=0.4] (xr0_1) at (6,1) {};
   \draw[<-,draw opacity=.7] (y2_1) -- node[left] {$\rmR_2$} ++ (xl1);
   \draw[<-,draw opacity=.7] (y2_1) -- node {$\rmR_1$} ++ (y1);
   \draw[<-,draw opacity=.7] (y2_1) -- node {$\rmR_{-2}$} ++ (xr0);
   
   \node[circle,draw,fill={lightblue}, opacity=0.4] (xl1_2) at (1,2) {};
   \node[circle,draw,fill={lightgreen}] (y1_2)  at (2,2) {};
   \node[circle,draw,fill={lightgreen}, opacity=0.4] (y2_2)  at (3,2) {};
   \node[circle,draw,fill={lightgreen}, opacity=0.4] (y3_2)  at (4,2) {};
   \node[circle,draw,fill={lightgreen}, opacity=0.4] (eoi_2)  at (5,2) {};
   \node[circle,draw,fill={lightblue}, opacity=0.4] (xr0_2) at (6,2) {};
   \draw[<-,draw opacity=.7] (y1_2) -- node {$\rmR_1$} ++ (xl1_1);
   \draw[<-,draw opacity=.7] (y1_2) -- node {$\rmR_{-2}$} ++ (xr0_1);
   
   \node (pd) at ($(y1_2) + (0, +0.8cm)$) {$\estinsertptok{\evy_{1}}{\evx_l}{\evx_r}$};
   \draw[->] (y1_2) -- (pd);
   
   \node[circle,draw,fill={lightpurple}] (r_xl1) at (7,0) {};
   \node at ($(r_xl1) + (0, -0.5cm)$) {$\evx_l$};
   
   \node[circle,draw,fill={lightpurple}] (r_y1) at (8,0) {};
   \node at ($(r_y1) + (0, -0.5cm)$) {$\evy_{1}$};
   \node[circle,draw,fill={lightpurple}] (r_y2) at (9,0) {};
   \node at ($(r_y2) + (0, -0.5cm)$) {$\evy_{2}$};
   \node[circle,draw,fill={lightpurple}] (r_eoi) at (10,0) {};
   \node at ($(r_eoi) + (0, -0.5cm)$) {$\eoi$};
   
   \node[circle,draw,fill={lightpurple}] (r_xr0) at (11,0) {};
   \node at ($(r_xr0) + (0, -0.5cm)$) {$\evx_r$};
   
   \node[circle,draw,fill={lightblue}, opacity=0.4] (r_xl1_1) at (7,1) {};
   \node[circle,draw,fill={lightgreen}, opacity=0.4] (r_y1_1)  at (8,1) {};
   \node[circle,draw,fill={lightgreen}] (r_y2_1)  at (9,1) {};
   \node[circle,draw,fill={lightgreen}, opacity=0.4] (r_eoi_1)  at (10,1) {};
   \node[circle,draw,fill={lightblue}, opacity=0.4] (r_xr0_1) at (11,1) {};
   \draw[<-,draw opacity=.7] (r_y2_1) -- node[left] {$\rmR_2$} ++ (r_xl1);
   \draw[<-,draw opacity=.7] (r_y2_1) -- node {$\rmR_1$} ++ (r_y1);
   \draw[<-,draw opacity=.7] (r_y2_1) -- node {$\rmR_{-2}$} ++ (r_xr0);
   
   \node[circle,draw,fill={lightblue}, opacity=0.4] (r_xl1_2) at (7,2) {};
   \node[circle,draw,fill={lightgreen}] (r_y1_2)  at (8,2) {};
   \node[circle,draw,fill={lightgreen}, opacity=0.4] (r_y2_2)  at (9,2) {};
   \node[circle,draw,fill={lightgreen}, opacity=0.4] (r_eoi_2)  at (10,2) {};
   \node[circle,draw,fill={lightblue}, opacity=0.4] (r_xr0_2) at (11,2) {};
   \draw[<-,draw opacity=.7] (r_y1_2) -- node {$\rmR_1$} ++ (r_xl1_1);
   \draw[<-,draw opacity=.7] (r_y1_2) -- node {$\rmR_{-2}$} ++ (r_xr0_1);
   
   \node (r_pd) at ($(r_y1_2) + (0, +0.8cm)$) {$\estinsertptok{\evy_{1}}{\evx_l}{\evx_r}$};
   \draw[->] (r_y1_2) -- (r_pd);
\end{tikzpicture}}
  \caption{\label{fig:xl_editor_attn} Attention in XL-Editor, the query stream (green nodes) are computed in the same way without any assumption of $|\vy|$.}
  \end{subfigure}
  \caption{\label{fig:attn_comp} Comparison between XLNet and XL-Editor for the relative positional encoding being used in attention.
  }
\end{figure}

To have a more illustrative example, consider estimating the probability of inserting a variable-length sequence $\vy$ between two tokens, $x_l$ and $x_r$:
\begin{align*}
  p_{\theta}(\vy|\evx_l,\evx_r) =\prod_i^{|\vy|} p_{\theta}(\evy_i|\evx_l,\vy_{<i},\evx_r).
\end{align*}
As shown in Figure~\ref{fig:xlnet_attn}, when XLNet predicts the probability of $\evy_k$, the relative positional encoding used between $\evy_k$ and $x_r$ would be $\rmR_{k-|\vy|-1}$, which makes the model aware of the length of the remaining tokens, $\vy_{k+1:|\vy|}$. Therefore, the trained probability estimations for XLNet is implicitly conditioned on the length of $\vy$, i.e., it is estimating $p_{\theta}(\vy|\evx_l,\evx_r, |\vy|)$.

We propose to remove the encoded length for the remaining tokens $\vy_{k+1:|\vy|}$ by treating the length of $\vy_{k+1:|\vy|}$ as 1 regardless of the actual length when selecting the relative positional encoding between the position of $y_k$ and other positions. The relative positional encoding between any other positions within the left and the right contexts are selected as if the length of the entire $\vy$ is 1. In addition, we train the model to predict a special $\eoi$ token at the end of the insertion so we could know when to stop insertion during inference. Effectively, the insertion-based relative encoding scheme eliminates the encoded distance between the left context and the right context but still indicates the exact position for insertion. This makes our method similar to a sequence-to-sequence model where both the left and right contexts are treated as the source sequence while the sequence being inserted is treated as the target sequence, but our model additionally encodes the relative positional relationships between the left context and the inserted sequence in a consistent manner.

More precisely, for the variable-length insertion probability estimation, $\insertp{\vy}{\vx_{1:i}}{\vx_{i+1:|\vx|}}$, let $\vz= \vx_{1:i}\oplus \vy\oplus\eoi\oplus\vx_{i+1:|\vx|}$, and we select the indices $a$, and $b$ so that $\vz_{a:b} = \vy\oplus\eoi$. Then, we replace the relative positional encoding $\rmR_{i-j}$ between the $i$-th and $j$-th position in $\vz$ with $\rmR_{i-j-\text{sign}(i-j)\phi_{a,b}(i, j)}$, where $\phi_{a, b}(\cdot)$ is defined as follows,
{
\begin{align*}
\fontsize{9.5pt}{10.5pt} \selectfont
\phi_{a,b}(i, j) = \begin{cases}
0, & \text{if both } i, j < a \text{ or both }  i, j > b\\
b-a, & i<a \text{ and } j > b \\
b-a, & i > b \text{ and } j < a \\
0, & a \le i \le b \text{ and } j \le i \\
b - i -1, & a \le i \le b < j\\
\text{unspecified}, & \text{otherwise.} \\
\end{cases}
\end{align*}
}

As shown in Figure~\ref{fig:xl_editor_attn}, the formulation ensures that the computed prediction on $\evy_k$ is equivalent regardless of the number of tokens in $\vy_{k+1:|\vy|}$. Thus the probability estimation no longer has implicit length assumption. Moreover, because the inserted tokens do not affect the computed hidden states, there is no need for re-computation in both training and inference phrase, overcoming the computation bottleneck existed in earlier works.
Note that the unspecified $(i, j)$ pairs are disallowed because the future positions, i.e., the remaining tokens  in $\vy_{k+1:|\vy|}$, cannot be attended to when predicting past tokens. While we only define $\phi(\cdot)$ with one sequence $\vz_{a:b}$ being inserted, it can be generalized to multiple inserted sequences in different positions by treating each one to have a length of 1, but we left further investigation to future works.

\subsection{Training Procedure for XL-Editor}
Equipped with proposed insertion-based relative positional encoding, we introduce our unsupervised training objective as follows, 

\begin{align}
\max_\theta \mathop{\E}_{\vx \sim \pdata} \mathop{\E}_{(i,j) \sim U(\vx)}&\estinsertp{\vx_{i:j}}{\vx_{1:i-1}}{\vx_{j+1:|\vx|}} \label{eq:xl_editor_obj}\\
=\max_\theta \mathop{\E}_{\vx \sim \pdata} \mathop{\E}_{(i,j) \sim U(\vx)}&[\prod_{t=i}^{j} \estinsertptok{x_t}{\vx_{1:t-1}}{\vx_{j+1:T}}\notag \\
& \times \estinsertptok{\eoi}{\vx_{1:j} }{\vx_{j+1:T}} \notag],
\end{align}
where $\vx$ is uniformly sampled from the training set, and $U(\vx)$ denotes a uniform random distribution among all non-empty intervals in $\vx$, i.e., $\{(i, j) | 1 \le i < j \le |\vx|\}$.

\subsection{Execute Post-editing Operations}

Next, we propose several possible strategies to execute post-editing operations with the probability estimation $q(\cdot)$.

\subsubsection{Locate} The position for insertion can be determined with $\argmin_i \estinsertptok{\eoi}{\vx_{1:i}}{\vx_{i+1:T}}$.

\subsubsection{Insert}
The probability of inserting $\vy$ between $\vx_{1:i}$ and $\vx_{i+1:T}$ can be estimated by $\estinsertp{\vy}{\vx_{1:i}}{\vx_{i+1:T}}$.

\subsubsection{Replace}
The odds of replacing a subsequence $\vx_{i:j}$ with a alternative sequence $\vy$ can be estimated by,
\begin{align*}
\frac{\estinsertp{\vy}{\vx_{1:i-1}}{\vx_{j+1:T}}}{\estinsertp{\vx_{i:j}}{\vx_{1:i-1}}{\vx_{j+1:T}}}.
\end{align*}

\subsubsection{Delete}
The deletion operation can be viewed as a special case of replacement where $\vy = \epsilon$ is an empty sequence.

\subsection{Post-editing Experiments}

To validate the effectiveness of our method, we conduct ablation studies for several post-editing tasks, including Locate (which position to insert), Text Infilling (what tokens to be inserted), and Text Deletion (what tokens to be deleted).

\subsubsection{CNN Dataset} We will use the CNN News articles collected by \citeauthor{hermann2015teaching}~\shortcite{hermann2015teaching} for our experiments. In particular, we sample 88,048 articles for training, and 2,000 articles for testing. There are approximately 40M  tokens in the train split and 900K tokens in the test split. 

\subsubsection{Baselines} We compare three models: (a) XL-Editor: the model is trained with the objective described in Equation (\ref{eq:xl_editor_obj}), where each sampled sequence $\vx$ consists of three consecutive sentences from the training articles. (b) $\text{XLNet}_\text{L2R}$: the model is trained with the same procedure except that insertion-based relative positional encoding is not used and no $\eoi$ was inserted. Therefore, for each sequence $\vx_{1:T}$ and randomly sampled interval $\vx_{i:j}$, we maximize $p_\theta(\vx_{i:j}|\vx_{1:i-1},\vx_{j+1:T})$ and factorize it in a left-to-right manner. While it is possible to use a randomly permuted factorization order, we find it to always perform worse if during inference we use a left-to-right factorization. Therefore, we omit the results of the permuted model. (c) Transformer:  the same objective as XL-Editor is optimized, but we set $\vx_{1:i-1} \oplus \masked\oplus \vx_{j+1:T}$ as the source sequence and $\vx_{i:j}$ as the target sequence, where the special token $\masked$ is used to denote the position for insertion, and train the transformer architecture \cite{vaswani2017attention} for this Seq2Seq problem.

Each model has 4 layers, 4 attention heads, and the hidden size is set as 256, while the inner hidden size of the positionwise feedforward network is set as 512. The models are randomly initialized and then trained with randomly sampled sequences from the training articles without specific design for each downstream task. Note that the transformer has more parameters due to the encoder-decoder architecture. In addition, we fine-tune the 12-layer $\text{XLNet}_{\text{BASE}}$ model \cite{yang2019xlnet} with $\text{XLNet}_\text{L2R}$ and XL-Editor objectives on training articles, and report their performance.

\subsubsection{Locate}
The artificial test set for the Locate task is constructed by sampling 5,000 sentences from the test split, and for each sentence, a randomly selected subsequence is deleted. The model is given 5 positions in the resulting sequence and is asked to detect which position to insert texts. For example, given the five positions in the following sentence, an insertion operation is required for the $4$-th position to make the sentence grammatical.

\begin{quote}
The $\red{\triangledown_1}$ state and EPA $\red{\triangledown_2}$ have found $\red{\triangledown_3}$ poor air quality $\red{\triangledown_4}$ San Joaquin $\red{\triangledown_5}$ Valley.
\end{quote}

For both XL-Editor and transformer, the prediction is made by selecting $\argmin_i \estinsertptok{\eoi}{\vx_{1:i}}{\vx_{i+1:T}}$. The XLNet does not offer a method for detecting insertion positions, so we use $\argmin_i p_\theta(\vx_{i:i+1}|\vx_{1:i-1},\vx_{i+2:T})$ as its prediction. The intuition is that if it is required to insert something between $\evx_i$ and $\evx_{i+1}$, then $\vx_{i:i+1}$ would have a low probability. The evaluation metrics is the accuracy for position prediction.

\subsubsection{Text Infilling}

In the earlier works of text infilling \cite{zhu2019text,liu-etal-2019-tigs}, the evaluation was mainly based on the fluency of the system outputs because there are many alternative ways to infill a given text, and there are no consistent criteria to determine which are more desirable.
We instead restrict the possibilities of plausible insertions by providing more context. In particular, 5,000 test instances are created, and each consists of three consecutive sentences, where the middle sentence has some random subsequence deleted. The model is asked to infill the deleted subsequence. The left and the right sentences restrict the search space of the insertion because the inserted text must make the paragraph coherent.

For both the transformer and XL-Editor, prediction is made by greedy decoding. However, there is no clear way for XLNet to determine the length of the inserted sequence. So we allow XLNet to enumerate all possible lengths up to $\text{max}(10, 2\times \text{truth length})$ and predict an inserted sequence for each length. The prediction $\vy$ that has the lowest perplexity for $p_\theta(\vy|\vx_{1:i}, \vx_{i+1:|\vx|})$ is selected as the output. In addition, we try a different ranking criterion by actually insert $\vy$ into the sequence and select the one with the lowest perplexity for the whole sequence $p_\theta(\vx_{1:i}\oplus \vy \oplus \vx_{i+1:|\vx|})$, this method is denoted as $\text{XLNet}_\text{L2R+ rank}$.

We evaluate the system outputs by computing the BLEU scores between the inserted sequence generated by each model against the originally deleted subsequence. \footnote{Note that this is different from earlier works in which the BLEU scores were computed against the whole completed sequences and they also used many sampled reference sequences so as to focus the evaluation on the fluency.}
A sample test instance and models' prediction are shown in Table~\ref{tab:infill_texts}.
Although none of the models can reproduce the deleted subsequence, the inserted output from the fine-tuned XL-Editor is more semantically coherent with the nearby context of the given paragraph compared to other baselines.

\begin{table}[ht]
  \centering
  \begin{tabular}{lp{5.2cm}}
    \toprule
    \multicolumn{2}{p{8.0cm}}{
   \textbf{Input:}  Until last year, Samoa and American Samoa  celebrated the new year on the same day. But then Samoa [\_] more easily with countries such as Australia and New Zealand. Because the date line is not fixed by any international law or agreement, it can zig and zag to accommodate such government and business interests.}\\ 
   \midrule
   \multicolumn{2}{p{8.0cm}}{\textbf{Answer:} hopped west of the line so it could trade}\\
   \midrule
   Transformer &'s " re-engagement " is \\
   $\text{XLNet}_\text{L2R+ rank}$ & 's decision to return to the islands is \\ 
   XL-Editor& , the country's largest economy, has \\
   \midrule
   \multicolumn{2}{l}{$\text{XLNet}_{\text{BASE}}$ fine-tuned as} \\
    $_\rightarrow\text{XLNet}_\text{L2R+ rank}$ &'s government and business interests began to clash \\
    $_\rightarrow$ XL-Editor& 's new date line has been changed to coincide \\
    \bottomrule
  \end{tabular}
  \caption{Infilled text from different models.}
  \label{tab:infill_texts}
\end{table}

\subsubsection{Text Deletion}
For text deletion, 5,000 test instances are created, and each consists of 5 consecutive sentences sampled from the testing articles. However, one of the 3 sentences in the middle was actually randomly sampled from a different article, and therefore needs to be deleted to make the text coherent. The model is asked to determine which sentence is the randomly inserted one.

For both transformer and XL-Editor, the prediction is determined by the perplexity ratio: 
\begin{align*}
\argmax_{i, j} \frac{
\text{PPL}(\estinsertp{\vx_{i:j}}{\vx_{1:i-1}}{\vx_{j+1:T}})}{
\text{PPL}(\estinsertptok{\eoi}{\vx_{1:i-1}}{\vx_{j+1:T}})
},
\end{align*}
while for XLNet, the prediction is made with,
\begin{align*}
\argmax_{i, j} \frac{
\text{PPL}(\estinsertp{\vx_{i-1:j+1}}{\vx_{1:i-2}}{\vx_{j+2:T}})}{
\text{PPL}(\estinsertp{\evx_{i-1}\oplus \evx_{j+1}}{\vx_{1:i-2}}{\vx_{j+2:T}})
}.
\end{align*}

In addition, we test a different ranking strategy by actually delete the subsequence and compute the perplexity of the whole sequence and select the one with lowest perplexity. This is denoted with + rank in the evaluation results.

\subsubsection{Discussions} The results are reported in Table~\ref{tab:postedit}. As we could see, XL-Editor has comparable or better performance in all three tasks. In addition, the best results are achieved by fine-tuning the $\text{XLNet}_\text{BASE}$ model with the XL-Editor objective, showing XL-Editor's cabability to leverage the pretrained XLNet to enable post-editing capabilities.

\begin{table}[ht]
  \centering
  \begin{tabular}{lccccc}
    \toprule
    \multirow{2}{*}{Model}& Locate &
    Infill & Delete \\
    \cmidrule(lr){2-2} \cmidrule(lr){3-3} \cmidrule(lr){4-4}
              &  ACC &  BLEU  & ACC     \\
   \midrule
   Random &  20.00 & - & 33.33 \\
   \midrule
   Transformer & 40.82 &  1.45 & 53.94  \\
   $\text{Transformer}_\text{+ rank}$ &-&  - & 39.70  \\
   $\text{XLNet}_\text{L2R}$ & 42.78 &0.49& 46.04 \\
   $\text{XLNet}_\text{L2R+ rank}$&- &\textbf{1.57} & 56.22\\
   XL-Editor & \textbf{50.48}&  \textbf{1.57} & 51.64\\
   $\text{XL-Editor}_\text{+ rank}$ & -&-& \textbf{57.44}  \\
   \midrule
     $\text{XLNet}_{\text{BASE}}$ fine-tuned as &  &  &  \\
     $_\rightarrow$ $\text{XLNet}_\text{L2R}$  & 47.56 & 1.29 & 54.10 \\
     $_\rightarrow$ $\text{XLNet}_\text{L2R+ rank}$ & - & 4.11  & 67.44 \\
     $_\rightarrow$ XL-Editor  & \textbf{56.32} & \textbf{4.37} &  \textbf{68.08}\\
     $_\rightarrow$ $\text{XL-Editor}_\text{+ rank}$  & -  & -  &65.76 \\
    \bottomrule
  \end{tabular}
  \caption{Evaluation results for post-editing on CNN dataset.}
  \label{tab:postedit}
\end{table}

\section{Unpaired Text Style Transfer}

Given $M$ predefined text styles, and a corpus $D=\{(\vx^{(i)}, \rs^{(i)})\}_{i=1}^N$ consisting of $N$ training instances, where $\vx^{(i)}$ is the $i$-th sequence while $\rs^{(i)} \in \{1, \ldots, M\}$ is the style of $\vx^{(i)}$.
The text style transfer task aims to learn a model that can transform any given sequence $\vx$ into a target sequence so that the style of the target sequence will become a desired target style $\rs_{\text{tgt}}$. Moreover, the target sequence should preserve as much semantic meaning from the original sequence as possible. For example, given a negative product review, we ask the model to transform it into a positive product review in which the same aspects of the same product are being discussed in a positive way. The task is challenging because there are usually no parallel training pairs available, and the models must be trained only with instance-wise style labels.

The style transfer task can be naturally viewed as a post-editing problem, where the editor aims to post-edit the source sequence to have the target style. Consider the conditional version of the insertion probability estimation,
\begin{align*}
\estinsertpcond{\vx_{i:j}}{\vx_{1:i-1}}{\vx_{j+1:T}}{, \rs},
\end{align*}
which gives the probability estimation of inserting $\vx_{i:j}$ for a given style $\rs$. Suppose we want to convert a sequence $\vx_{1:T}$ from style $\rs_{\text{src}}$ to $\rs_{\text{tgt}}$, we propose to estimate the odds that a subsequence $\vx_{i:j}$ needs replacement by,

\begin{align*}
f_\theta(i, j|\vx_{1:T},\rs_\text{src}\rightarrow \rs_\text{tgt})=\frac{\estinsertpcond{\vx_{i:j}}{\vx_{1:i-1}}{\vx_{j+1:T}}{, \rs_{\text{src}}}}{\estinsertpcond{\vx_{i:j}}{\vx_{1:i-1}}{\vx_{j+1:T}}{, \rs_{\text{tgt}}}}.
\end{align*}

The intuition is that if a subsequence has high probability for style $\rs_{\text{src}}$ but low probability for style $\rs_{\text{tgt}}$, then it probably needs to be replaced in order to transform the sequence into style $\rs_{\text{tgt}}$. Once we identify the place for replacement, we could replace it with a sequence $\vy$ by sampling from $\estinsertpcond{\vy}{\vx_{1:i-1}}{\vx_{j+1:T}}{, \rs_{\text{tgt}}}$. In addition, we can also identify the location for insertion in a similar way,

\begin{align*}
f_\theta(i, i-1|\vx_{1:T},\rs_\text{src}\rightarrow \rs_\text{tgt})=\frac{\estinsertpcond{\epsilon}{\vx_{1:i-1}}{\vx_{i:T}}{, \rs_{\text{src}}}}{\estinsertpcond{\epsilon}{\vx_{1:i-1}}{\vx_{i:T}}{, \rs_{\text{tgt}}}},
\end{align*}
where $\vx_{i:i-1}$ is treated as an empty subsquence $\epsilon$.
The intuition is that if $\epsilon$ has high probability for $\rs_{\text{src}}$ but low probability for style $\rs_{\text{tgt}}$, then it probably means something can be inserted to transform the sequence into $\rs_{\text{tgt}}$. The detailed post-editing process is outlined in Algorithm~\ref{alg:sty}. Note the post-editing process allows three operations: (a) insertion is executed when $j = i-1$, (b) deletion is executed when $j \ge i$ and $\vy = \epsilon$, (c) replacement is executed when $j \ge i$ and $\vy \ne \epsilon$. When insertion is executed, we disallow the model to insert an empty $\vy$.

\begin{algorithm}
\caption{Post-editing for Style Transfer}
\label{alg:sty}
\begin{algorithmic}[1]
   \STATE {\bfseries Input:} Input sequence $\vx$, source style $\rs_\text{src}$, target style $\rs_\text{tgt}$, hyper-parameters $L, \rv_\text{thres}$.
   \WHILE{true}
   \STATE $C \leftarrow \{(a,b)| 1\le a \le |\vx|+1, 1\le b \le |\vx|$, \\
   $ \quad \quad \quad \quad \quad -1 \le b-a \le L\}$
   \STATE $ (i, j) \leftarrow \argmax_{(a,b) \in C}f_\theta(a, b|\vx,\rs_\text{src}\rightarrow \rs_\text{tgt})$
   \STATE $\rv_\text{max} \leftarrow f_\theta({i}, {j}|\vx,\rs_\text{src}\rightarrow \rs_\text{tgt})$
   \IF{$\rv_\text{max} \ge \rv_\text{thres}$}
   \STATE $\vy \leftarrow \argmax_\vy \estinsertpcond{\vy}{\vx_{1:{i-1}}}{\vx_{j+1:|\vx|}}{, \rs_{\text{tgt}}}$
   \STATE $\vx \leftarrow \vx_{1:i-1}\oplus \vy \oplus \vx_{j+1:|\vx|}$
   \ELSE
    \STATE Return the current $\vx$.
   \ENDIF
   \ENDWHILE
\end{algorithmic}
\end{algorithm}

\subsubsection{Training Objective}

Our XL-Editor is trained to optimize two objectives,
\begin{align}
\max_\theta \mathop{\E}_{(\vx, \rs) \sim \pdata} \mathop{\E}_{(i,j) \sim U(\vx)}\estinsertp{\vx_{i:j}}{\vx_{1:i-1}}{\vx_{j+1:|\vx|},\rs} \label{eq:xl_editor_sty_obj}
\end{align}
and 
\begin{align}
\max_\theta \mathop{\E}_{(\vx, \rs) \sim \pdata} p_\theta(\rs|\vx),\label{eq:xl_editor_aux_obj}
\end{align}
where (\ref{eq:xl_editor_sty_obj}) is the conditional version of the variable-length insertion probability estimation, which is implemented by appending a special style token representing $\rs$ to the sequence while estimating the probability, and (\ref{eq:xl_editor_aux_obj}) is auxiliary style prediction. We follow the  implementation of XLNet \cite{yang2019xlnet} by appending a $\cls$ token to $\vx$ and feed the top-layer hidden vector of $\cls$ to a feed-forward network for prediction.

\subsubsection{Other Task-specific Rules}
We further introduce two task specific tricks which are found to be helpful in our experiments -- (a) \textbf{biased sampling}: $\vy$ is sampled with greedy decoding, but when the first token is sampled, we select $y_1 = \argmax_{y^*} \estinsertptok{y^*}{\vx_{1:{i-1}}}{\vx_{j+1:|\vx|},\rs_\text{tgt}} - \estinsertptok{y^*}{\vx_{1:{i-1}}}{\vx_{j+1:|\vx|},\rs_\text{src}}$ so as to increase the intensity of the style, and (b) \textbf{forced insertion}: for the Yelp dataset used in our experiments, there exist many incomplete sentences. Therefore, we force XL-Editor to make an insertion at the start of the sentence when no other operations can be executed by Algorithm~\ref{alg:sty} and $p_\theta(\rs_\text{src}|\vx)$ is greater than $0.9$. We found that this allows the model to complete the sentence in a way to enforce the target style without distorting the sentence. Forced insertion is not applied to the Amazon dataset.

\subsection{Style Transfer Experiments}

We conduct text style transfer experiments on two widely used datasets in text style transfer, the Yelp and Amazon datasets~\cite{li2018transfer}. Each of these datasets contains two styles: positive and negative sentiment, and the task is to flip the sentiment of a given review text. The dataset statistics are shown in Table~\ref{tab:dataset_stat}. We directly re-use the same 4-layer architecture as in Section~\ref{sec:xl_editor} and train XL-Editor from scratch on the training split. We do not tune any training parameters, and since reference texts are not available in the dev split, we tune the inference parameters by observing the model behaviour. The same parameters tuned for each dataset are directly applied for post-editing different system outputs.

\begin{table}[]
  \centering
  \begin{tabular}{lcccc}
    \toprule
    Dataset& Style &  Train & Dev & Test  \\
   \midrule
   \multirow{2}{*}{Yelp} & Positive &  266K & 2000 & 500 \\
    & Negative &  177K & 2000 & 500\\
   \multirow{2}{*}{Amazon} & Positive &  277K & 985 & 500 \\
    & Negative &  278K & 1015 & 500\\
    \bottomrule
  \end{tabular}
  \caption{Datset statistics.}
  \label{tab:dataset_stat}
\end{table}

\subsubsection{Baselines}

The baselines we compare with can be roughly categorized into 4 groups:
(a) Rule-based systems that delete and infill sentiment words for the given sentences, including \textbf{DeleteOnly (Del)}, \textbf{TemplateBased (Tpl)}, and 
\textbf{Del-Ret-Gen (DRG)} \cite{li2018transfer};
(b) Disentangled representation learning systems that generate outputs from the learned representations, including \textbf{CrossAligned (XAli)} \cite{shen2017style}, \textbf{MultiDecoder (MDec)}, \textbf{StyleEmbedding (SEmb)} \cite{fu2018style}, and \textbf{BackTranslate (BT)} \cite{prabhumoye2018style};
(c) Sequence-to-sequence systems that generate target sequences without explicit disentangled representations, including \textbf{UnsuperMT (UMT)} \cite{zhang2018style} and  \textbf{Style Transformer Conditional/Multi-class (StyTc/StyTm)}~\cite{dai2019style};
(d) Reinforcement learning systems that use cycle-consistency constraints to train agents for transforming the source sequence into the target style, including \textbf{UnpairedRL (URL)} \cite{xu2018unpaired}, \textbf{DualRL (DRL)} \cite{Luo19DualRL}, and  \textbf{Point-then-Operate (PTO)} \cite{WuRLS19}.
In addition, we report the results for directly outputting the input sequence without modification~\textbf{(Copy)}.

\subsubsection{Evaluation Results} We use the standard BLEU score computed against the human reference texts, the style accuracy produced by a style classifier, and the G-score as used in \cite{xu2018unpaired}, which is the geometric mean between BLEU and style accuracy. We adopt the evaluation code from \citeauthor{WuRLS19}~\shortcite{WuRLS19}. We obtain the system outputs of StyTc, STyTm, DRL, UMT from the authors for evaluation. The evaluation results for other systems are directly reproduced from \citeauthor{WuRLS19}~\shortcite{WuRLS19}. We apply our \textbf{XL-Editor (XLE)} to post-edit the system outputs of the two best-performing systems UMT and PTO. The results are shown in Table~\ref{tab:auto_result}. In most cases, our post-editing procedure achieves significant improvement in style accuracy while making little sacrifice in BLEU. The best G-score are achieved by our systems in both datasets.
\begin{table}[ht]
  \centering
  \begin{tabular}{lcccccc}
    \toprule
    \multirow{2}{*}{Model}& \multicolumn{3}{c}{Yelp} &  \multicolumn{3}{c}{Amazon} \\
    \cmidrule(lr){2-4} \cmidrule(lr){5-7}
         & Acc   & Bleu &  G &  Acc    & Bleu  & G  \\
   \midrule
    Copy           & 2.4  & 32.39  &8.8 & 17.2 & \textbf{48.38} & 28.8 \\
    \midrule
    XAli         & 74.7  & 9.06 & 26.0&  75.1 & 1.90 & 11.9  \\
    MDec         & 50.6  & 14.54& 27.1&  69.9 & 9.07 & 25.2 \\
    SEmb       & 8.4  & 21.06& 13.3&  38.2 & 15.07 & 24.0 \\
    Tpl        & 81.2  & 22.57& 42.8&  64.3 & 34.79 & 47.3 \\
    Del           & 86.0& 14.64& 35.5& 47.0& 33.00& 39.4  \\
    DRG          & 88.6& 15.96& 37.6& 51.0& 30.09& 39.2  \\
    BT        & 94.6& 2.46&  15.3& \textbf{76.7}& 1.04& 8.9  \\
    URL           & 57.5& 18.81&  32.9& 56.3& 15.93& 29.9  \\
    DRL               & 89.0 &27.91 &  49.8 & N/A &N/A & N/A \\
    StyTc   & 91.1& 24.57& 47.3 &N/A  &N/A  &N/A \\
    StyTm   &86.0& 28.27 & 49.3& N/A & N/A & N/A \\  \midrule
    UMT           & 97.9& 22.68& 47.1& 72.4& 33.34& 49.1  \\
    $+\text{XLE}$ & \textbf{98.8} &22.37 & 47.0&72.9 &33.27 &\textbf{49.2} \\
    \midrule
    PTO & 91.5& \textbf{29.86}& 52.3&  40.2 & 41.86  & 41.0\\
    $+\text{XLE}$ &95.7 &28.75 &\textbf{52.5} &43.2 & 41.13& 42.2\\

    \bottomrule
  \end{tabular}
  \caption{Evaluation results on Yelp and Amazon datsets.}
  \label{tab:auto_result}
\end{table}

\subsubsection{Discussions}

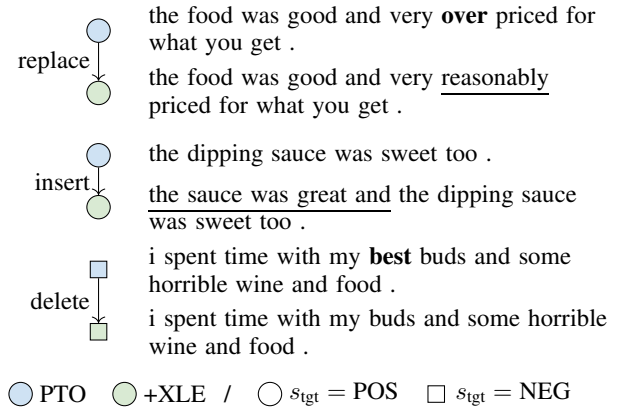
\begin{figure}[ht]
  \resizebox{.99\columnwidth}!{ 
  \begin{tikzpicture}[scale=0.7,every node/.style={scale=0.95}]
\tikzstyle{nome}=[rectangle,anchor=west]
\node[circle,draw] (neg2pos) at (4.3,2) {};
\node[rectangle,right] (neg2pos_text) at ($(neg2pos) + (0.2cm, 0)$) {$s_\text{tgt} = \text{POS}$};

\node[rectangle,draw] (pos2neg) at (7.5,2) {};
\node[rectangle,right] (pos2neg_text) at ($(pos2neg) + (0.2cm, 0)$) {$s_\text{tgt} = \text{NEG}$};

\node[circle,draw,fill={lightblue}] (pto) at (-0.5,2) {};
\node[rectangle,right] (pto_text) at ($(pto) + (0.2cm, 0)$) {PTO};
\node[circle,draw,fill={lightgreen}] (xle) at (1.5,2) {};
\node[rectangle,right] (xle_text) at ($(xle) + (0.2cm, 0)$) {+XLE{ }{ }{ }/};

\node[circle,draw,fill={lightblue}] (yelp_n2p_1_p) at (1,9) {};
\node[rectangle,right,text width=6.5cm] (yelp_n2p_1_p_text) at ($(yelp_n2p_1_p) + (0.8cm, 0)$) {the food was good and very \textbf{over} priced for what you get .};
\node[circle,draw,fill={lightgreen}] (yelp_n2p_1_x) at (1,7.8) {};
\node[rectangle,right,text width=6.5cm] (yelp_n2p_1_x_text) at ($(yelp_n2p_1_x) + (0.8cm, 0)$) {the food was good and very \underline{reasonably} priced for what you get .};
\draw[->] (yelp_n2p_1_p) -- node[left] {replace}  ++ (yelp_n2p_1_x);

\node[circle,draw,fill={lightblue}] (yelp_n2p_2_p) at (1,6.6) {};
\node[rectangle,right,text width=6.5cm] (yelp_n2p_2_p_text) at ($(yelp_n2p_2_p) + (0.8cm, 0)$) {the dipping sauce was sweet too . };
\node[circle,draw,fill={lightgreen}] (yelp_n2p_2_x) at (1,5.6) {};
\node[rectangle,right,text width=6.5cm] (yelp_n2p_2_x_text) at ($(yelp_n2p_2_x) + (0.8cm, 0)$) {\underline{the sauce was great and} the dipping sauce was sweet too .};
\draw[->] (yelp_n2p_2_p) -- node[left] {insert}  ++ (yelp_n2p_2_x);

\node[rectangle,draw,fill={lightblue}] (yelp_p2n_1_p) at (1,4.4) {};
\node[rectangle,right,text width=6.5cm] (yelp_p2n_1_p_text) at ($(yelp_p2n_1_p) + (0.8cm, 0)$) {i spent time with my \textbf{best} buds and some horrible wine and food .};
\node[rectangle,draw,fill={lightgreen}] (yelp_p2n_1_x) at (1,3.2) {};
\node[rectangle,right,text width=6.5cm] (yelp_p2n_1_x_text) at ($(yelp_p2n_1_x) + (0.8cm, 0)$) {i spent time with my buds and some horrible wine and food .};
\draw[->] (yelp_p2n_1_p) -- node[left] {delete}  ++ (yelp_p2n_1_x);

\end{tikzpicture}
  }
  \caption{\label{fig:xl_editor_op} Sample post-editing operations  on Yelp dataset.
  }
\end{figure}

We examined the post-editing operations executed by the XL-Editor. As shown in Figure~\ref{fig:xl_editor_op}, we find that XL-Editor is quite competent in picking up the small errors made by PTO. While our method is trained with standard maximum likelihood estimation on the inserted sequence, it is also possible to incorporate other well-explored techniques such as adversarial loss for text style transfer, which we leave for the future works.

\section{Conclusion and Future Works}
In this paper, we propose XL-Editor, a novel training framework that accommodates current state-of-the-art language pretraining model, XLNet, to support variable-length insertion probability estimation.
Armed with the novel insertion-based relative positional encoding, XL-Editor can not only efficiently estimate the insertion probability but also elegantly support several post-editing operations (i.e., insertion, deletion, and replacement) that is complementary to any existing sequence-to-sequence models.
We demonstrate the effectiveness of XL-Editor on the post-editing tasks as well as the text style transfer task. By post-editing the transferred sentences with XL-Editor, we observe clear improvement on style accuracy with little sacrifice in BLEU. Currently, our work only focuses on basic text editing operations and text style transfer. For the future work we will investigate other applications for our method.

\fontsize{9.0pt}{10.0pt} \selectfont
\bibliography{reference}
\bibliographystyle{aaai}
\end{document}